# FOMTrace: Interactive Video Segmentation By Image Graphs and Fuzzy Object Models

Thiago Vallin Spina and Alexandre Xavier Falcão

***Abstract*—Common users have changed from mere consumers to active producers of multimedia data content. Video editing plays an important role in this scenario, calling for simple segmentation tools that can handle fast-moving and deformable video objects with possible occlusions, color similarities with the background, among other challenges. We present an interactive video segmentation method, named *FOMTrace*, which addresses the problem in an effective and efficient way. From a user-provided object mask in a first frame, the method performs semi-automatic video segmentation on a spatiotemporal superpixel-graph, and then estimates a *Fuzzy Object Model* (FOM), which refines segmentation of the second frame by constraining delineation on a pixel-graph within a region where the object's boundary is expected to be. The user can correct/accept the refined object mask in the second frame, which is then similarly used to improve the spatiotemporal video segmentation of the remaining frames. Both steps are repeated alternately, within interactive response times, until the segmentation refinement of the final frame is accepted by the user. Extensive experiments demonstrate FOMTrace's ability for tracing objects in comparison with state-of-the-art approaches for interactive video segmentation, supervised, and unsupervised object tracking.**

## I. Introduction

Several hours of video footage with diversified content are uploaded every minute to many websites. This content is often represented by objects that can be isolated from the surrounding background for data analysis and/or video editing (e.g., alpha matting [1]). In video editing, accurate object segmentation is mandatory, albeit it is a time-consuming and error-prone task when done manually, and fully automatic solutions are inviable in general settings. This scenario calls for interactive video segmentation tools.

In interactive image segmentation, the human's superior ability for locating the object can be combined with the computer's advanced capacity for precise delineation in a synergistic way [2]. The user can use the knowledge about the object's appearance and location to provide sparse annotation (e.g., scribbles [3]–[7]), while the computer predicts the labels of the remaining pixels. When delineation errors occur, the user can add annotation to guide the computer towards correcting the result. We follow the same well succeeded principle for interactive video segmentation.

In this context, the user's time and involvement must be minimized, while preserving the user's control over the segmentation process with maximum accuracy [8]. Despite the recent impressive progress in interactive video segmentation [8]–[14], those goals remain a main issue due to object deformation, fast motion, occlusion, color similarity with the background, among other challenges. In view of that, we present an interactive video segmentation method, named FOMTrace, to address the problem effectively and efficiently.

From an input video $\mathbf{I}$ with $n_f$ frames, the user provides an object mask $L_0$ (label image) for the first frame $I_{t=0}$ by using standard interactive image segmentation techniques [7], [15], [16]. FOMTrace then interprets the video volume as a spatiotemporal superpixel-graph and uses the label image $L_0$ to propagate segmentation to the remaining frames $I_{t>0}$, automatically. A *Fuzzy Object Model* (FOM) is estimated and used to refine segmentation on a pixel-graph of the second frame. The user can correct/accept the refined object mask, which is then similarly used to improve the spatiotemporal video segmentation of the remaining frames. This process repeats with possible user supervision in a frame-by-frame fashion. Figure 1 illustrates the general pipeline of FOMTrace, which can be better described as follows.

At any time $t = 1, 2, \dots, n_f - 1$, the user can correct/accept a tentative object mask $L'_{t-1}$ for the previous frame, producing a final label image $L_{t-1}$. FOMTrace uses $L_{t-1}$ as input to predict the label images $\hat{L}_t, \hat{L}_{t+1}, \dots, \hat{L}_{n_f-1}$ of the remaining frames. At this point, the predicted label image $\hat{L}_t$ provides an approximation for the object's segmentation in the current frame $I_t$, but may contain errors (e.g., Figure 2c). For tentative correction, FOMTrace estimates a Fuzzy Object Model $O_t$ from the predicted and past label images $\hat{L}_t$ and $L_{t-1}$, respectively (Figures 2b-2d), which constrains delineation on a pixel-graph of the current frame $I_t$ within a region where the object boundary is expected to be. The refined object mask $L'_t$ (Figure 2e) is then displayed to the user, who may correct/accept it, creating the final label image $L_t$. The object mask $L_t$ becomes the input for video segmentation of the next frames and the process repeats until the user approves $L_{n_f-1}$.

It is important to note that the automatic video segmentation is repeated at every iteration to better deal with fast-moving deformable objects. A drawback is that errors occured in future frames, not being currently viewed by the user, may be brought back to the present, which is counter-intuitive [9] (e.g.,

This work was supported in part by the São Paulo Research Foundation under Grant 2011/01434-9 and in part by the National Council for Scientific and Technological Development under Grant 302970/2014-2 and Grant 479070/2013-0. The authors would like to thank the creators of the SegTrack and GeorgiaTech segmentation datasets for the videos.

T. V. Spina and A. X. Falcão are with the Institute of Computing of the University of Campinas, Campinas, SP, 13083-852 Brazil (email: {tvspina,afalcao}@ic.unicamp.br).

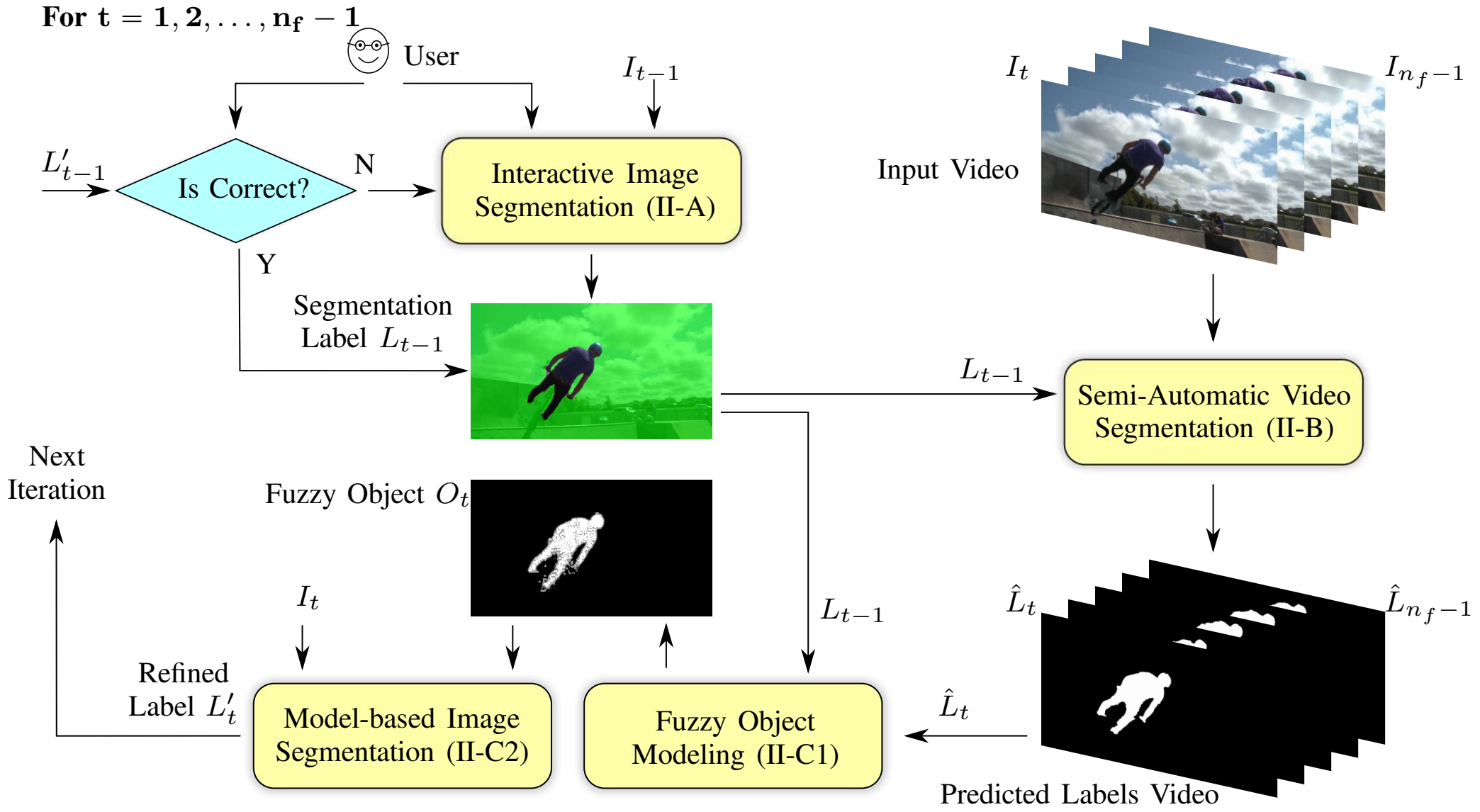

Fig. 1: Overall scheme for our interactive video segmentation method named *FOMTrace*. The diagram depicts the segmentation process for the current frame $I_t$, where $t > 0$, after FOMTrace was interactively initialized with a label mask $L_0$ in frame $I_0$. The numbers inside parentheses indicate the main sections in this paper where the corresponding block is detailed.

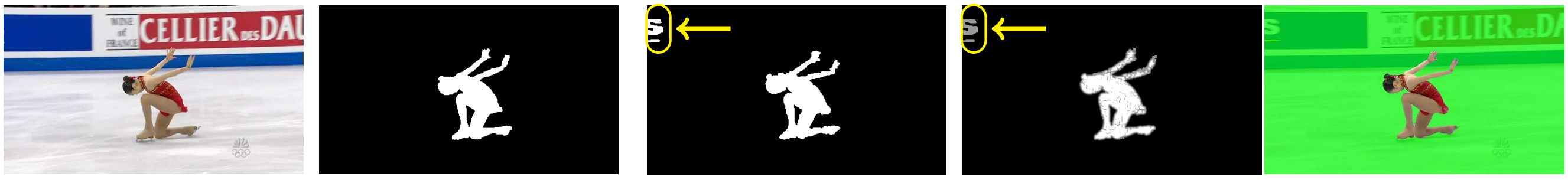

(a) Previous frame $I_{t-1}$. (b) Previous label $L_{t-1}$. (c) Predicted label $\hat{L}_t$ for frame $I_t$ with error (red arrow). (d) Fuzzy object model $O_t$. Gray regions indicate uncertainty. (e) Final segmentation for frame $I_t$.

Fig. 2: Automatic segmentation refinement by using a fuzzy object model for the current frame $I_t$. The previous label mask $L_{t-1}$ (b) is combined with the predicted label $\hat{L}_t$ (c) to form an object model represented by a fuzzy image $O_t$ (d), which is applied to correct segmentation in frame $I_t$ (e). The segmentation error is highlighted in yellow.

the error highlighted at the top row of Figure 3). Moreover, when the object shares similar colors with background regions that touch it, the video segmentation alone cannot properly segregate them. The fuzzy object model mitigates the problem, by taking into account the object's past and future contours as shape constraints to fix (refine) segmentation preemptively. In other words, the fuzzy image $O_t$ simulates the user's knowledge about the object's silhouette *evolution* across time, aiming to correct segmentation with no user intervention (i.e., no intervention is required at the bottom row of Figure 3). The refined segmentation is expected to better approximate the object's real shape in the current frame $I_t$. Therefore, it is used to improve video segmentation at each iteration.

For object delineation, the currect frame (image) and the video are interpreted as weighted graphs. In the former case, the pixels are the nodes and the arcs connect their 8-neighbors. In the latter case, the nodes are superpixels computed for each video frame and connected by using spatiotemporal arcs that go backwards and forwards in time between adjacent frames. For both types of graphs, delineation relies on optimum competition among internal and external nodes, called *seeds*, for their most closely connected nodes in the graph. The image graph is then partitioned into an optimum-path forest and the object is defined by trees rooted at its internal seeds [16], [17]. The seeds are automatically selected from the previous object mask for video segmentation and from the fuzzy object model for the segmentation refinement of the current frame.

The automatic video segmentation extends an interactive image segmentation method based on superpixel-graphs [17], being our first contribution in this paper. This extension is crucial in FOMTrace, given that it allows interactive response times to be reapplied on the volumetric graph of superpixels at every iteration, while producing accurate results for deformable objects. Indeed, if the automatic video segmentation from the first frame causes errors on a few frames (top row of Figure 3), as variant, the user can decide to make all corrections, ending the process in a single-shot.

Fuzzy object modeling is a recent trend in medical image segmentation [18]–[22], being a considerable more efficient alternative to statistical atlases [23], [24]. In medical imaging, the fuzzy object models are built from several segmentation masks of a desired object, with no need for deformable image registration. For segmentation of a new image, the model searches for the object’s location by using optimization techniques and search region information learned from training. We are extending these ideas to deformable objects in video, being a second important contribution of this work. We demonstrate in our experiments that our fuzzy object model in FOMTrace actually decreases the need for interactive corrections (Figure 3).

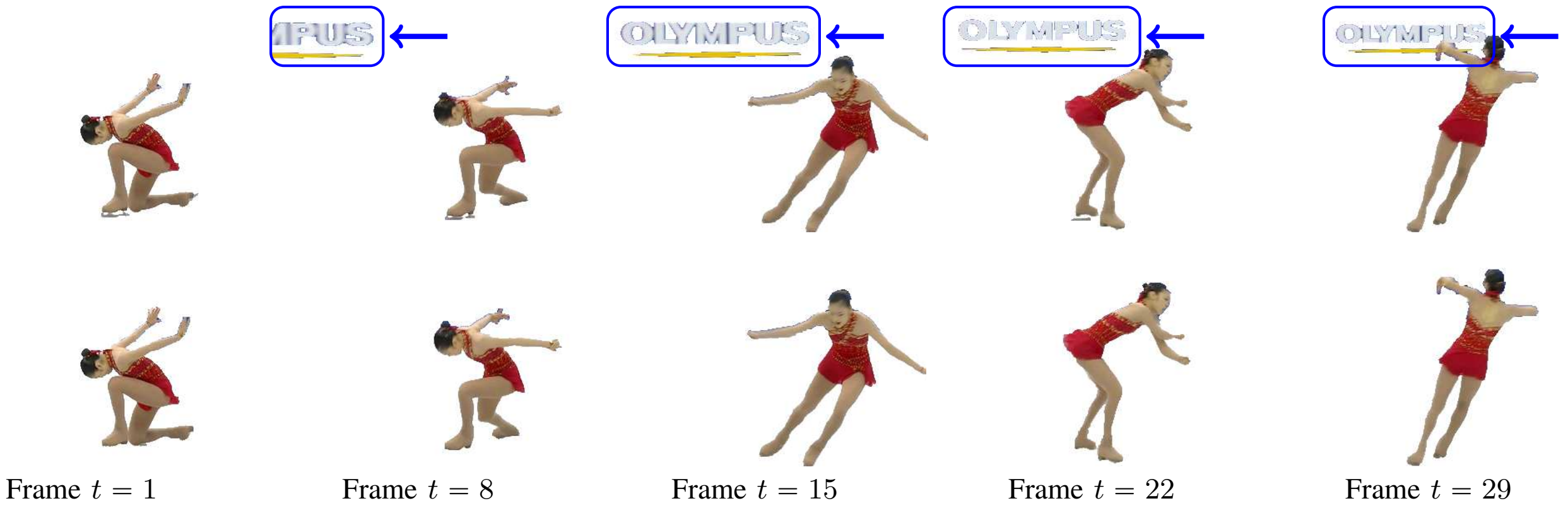

Fig. 3: FOMTrace without fuzzy object modeling (top row) and with it (bottom row). Due to our volumetric video segmentation step, a “leaking” that happened in a future frame (the “Olympus” sign at time $t = 29$, highlighted in blue) appears in all previous frames. The fuzzy object model can correct these errors in all intermediate frames. In both cases, the user only provides the label $L_0$.

Given that FOMTrace is related to several works in both video and medical image segmentation, for the sake of clarity we make the proper references to previous works at the end of some sections. Section II details the FOMTrace algorithm and, in Section III, experimental results in comparison with state-of-the-art approaches are presented and discussed. We finally provide concluding remarks, information about the remaining challenges and future work in Section IV.

## II. The FOMTrace Video Segmentation Method

In FOMTrace, video segmentation, model-based object refinement, and interactive segmentation correction are all based on the same object delineation algorithm derived from the Optimum-Path Forest (OPF) methodology. The OPF was first proposed for the design of image processing operators, being called *Image Foresting Transform* [3] (IFT), and subsequently extended to clustering and classification [25], [26].

In the OPF methodology, data samples (e.g., pixels, regions, images, objects) are interpreted as nodes of a graph $(\mathcal{N}, \mathcal{A})$, whose arcs connect samples according to some *adjacency relation* $\mathcal{A} \subseteq \mathcal{N} \times \mathcal{N}$. For a given problem, which can be directly or indirectly related to an optimum set partitioning problem, a suitable *connectivity function* assigns a value $f(\pi_q)$ to any path $\pi_q$ with terminus $q$ in the graph, including the trivial ones $\pi_q = \langle q \rangle$. From an initial connectivity map, where all paths are trivial, the algorithm propagates data properties in a non-decreasing order of path value, from the minima of the trivial map to the remaining nodes, such that each node is conquered by the minimum that offers an optimum path to it.

The OPF algorithm essentially minimizes a final connectivity map $C(q) = \min_{\forall \pi_q \in \Pi_q}\{f(\pi_q)\}$, by considering all possible paths $\pi_q$ in the set $\Pi_q$ of paths with terminus $q$, and outputs an *optimum-path forest* rooted at the minima of the map $C$ — i.e., an acyclic predecessor map $P$ that assigns to each node $q$ its predecessor $P(q) = p$ in the optimum path or a distinct marker $P(q) = nil \notin \mathcal{N}$, when the node $q$ is a root of the map. Other attributes, such as root $R(q)$ and root label $L(q)$, can also be assigned to each node $q \in \mathcal{N}$, and the problem is reduced to a local processing of these attributes.

For simplicity, we will define FOMTrace for the binary case, although it works for multiple objects. FOMTrace uses the IFT by *optimum seed competition* [3] (IFT-SC) for object delineation in image and video. In IFT-SC, seeds inside and outside the object compete among themselves and the object is defined by the optimum-path trees rooted at its internal seeds. We use IFT-SC on pixel- and superpixel-graphs. The three aforementioned operations in FOMTrace differ by the way the seeds are estimated, the type of graph, and the image properties used in the connectivity function of IFT-SC. These aspects are explained in the next sections, being first instantiated for pixel-graphs in interactive image (frame) segmentation.

**Related works:** In [27], IFT-SC was proven to be one out of two basic algorithms for image segmentation based on optimum graph cuts, the other being the min-cut/max-flow algorithm [4]. Superpixel-graphs have also been used for interactive video segmentation using min-cut/max-flow [28]. However, we chose IFT-SC because: i) it is more robust to seed positioning than min-cut/max-flow [29]; ii) it can execute in linear time, independently of the number of objects [3]; and iii) it allows segmentation corrections in sublinear time [16], even when combined with other IFT-derived methods [7].

### A. *Interactive Image Segmentation using IFT-SC*

An image $I : D_I \rightarrow \mathbb{V}$ is a mapping, where $D_I \subset Z^2$ is the *image domain* and $\mathbb{V} \subset \Re^m$ is the color space ($m = 1$ for monochromatic images and $m = 3$ for color images). That is, each pixel $p \in D_I$ is assigned a color $I(p)$. We represent images in the YCbCr color space.

For image segmentation, the node set $\mathcal{N} = D_I$ and the arcs $(p, q) \in \mathcal{A}$ are defined between 8-adjacent pixels. The idea is to draw markers (scribbles) inside and outside the object and define a connectivity function that penalizes paths from these markers when they try to cross the object's border during the algorithm. Each arc is then assigned a weight $w(p, q) \geqslant 0$, preferably higher on the object's border than elsewhere, and the connectivity function $f_{\max}$ forces the origin of the paths to be in the marker set $\mathcal{M}$ (seed nodes):

$$
\begin{aligned}
f_{\max}(\langle q \rangle) &= \begin{cases} 0 & \text{if } q \in \mathcal{M} \subset \mathcal{N}, \\ +\infty & \text{otherwise}, \end{cases} \\
f_{\max}(\pi_p \cdot \langle p, q \rangle) &= \max\{f_{\max}(\pi_p), w(p, q)\}, \qquad (1)
\end{aligned}
$$

where $\pi_p \cdot \langle p, q \rangle$ indicates the extension of a path $\pi_p$ by an arc $(p, q)$ and $\mathcal{M} \subset \mathcal{N}$ contains pixels of the interior and exterior of the object of interest in some frame $t$.

The simplest definition for the arc weights $w(p, q)$ is the image's gradient magnitude $w(p, q) = \|\nabla I(q)\|$, which makes it equivalent to the seeded watershed transform [30]. We have also used more elaborate functions based on intelligent interpretation of the user's input [6] for interactive segmentation.

The seed nodes in $\mathcal{M}$ then compete among themselves for their most closely connected nodes in the graph. Each node $p \in \mathcal{M}$ is assigned its true label $\lambda(p) \in \{0, 1\}$ (background or foreground) and each node $q \notin \mathcal{M}$ is assigned the label $L(q) \leftarrow \lambda(R(q))$ of its root node $R(q) \in \mathcal{M}$. The object can be directly obtained from the label map $L$.

Figure 4 presents an example of interactive segmentation using IFT-SC, whose process is explained with a fictious 4-neighborhood graph (top row). In this graph, two seed nodes in bold represent the user-drawn scribbles and bold arcs indicate higher weights between object and background. The optimum-path forest $P$ for $f_{\max}$ is shown in the bottom row.

### B. Semi-Automatic Video Segmentation

A video is a sequence of images $\mathbf{I} = \langle I_0, I_1, \ldots, I_{n_f-1} \rangle$ along time $0 \leqslant t \leqslant n_f - 1$. For video segmentation, we first segment each image $I_t$ into disjoint and 8-connected superpixels $\mathcal{S}_t^i$, $i = 1, 2, \ldots, n$, such that $\bigcup_{i=1}^{n} \mathcal{S}_t^i = \mathcal{S}_t$, and the superpixels $p = \mathcal{S}_t^i$ from all frames $t$ are used to compose the node set $\mathcal{N} = \bigcup_{t=0}^{n_f-1} \mathcal{S}_t$. *Symmetric* and *non-reflexive* arcs $(p, q) \in \mathcal{A}$ are defined by pairs of superpixels that: (i) share pixel edges in $I_t$, $t = 0, 1, \ldots, n_f - 1$ or (ii) present intersection $p \bigcap q \neq \varnothing$ given their pixel coordinates in two subsequent frames $I_t$ and $I_{t+1}$, before and after applying dense optical flow [31], similarly to [32] (Figure 5).

The superpixels in $\mathcal{S}_t$ for each frame $I_t$ are generated by using the Simple Linear Iterative Clustering algorithm [33] (SLIC). The arc weights $w(p, q) = \|\vec{v}(p) - \vec{v}(q)\|_2$ for the superpixel-graph consider the squared Euclidean distance between the feature vectors of superpixels $p, q \in \mathcal{S}$, where $\vec{v}(p)$ and $\vec{v}(q)$ are the mean colors of the underlying pixels [17].

At any given time $t > 0$, the accepted pixel label image $L_{t-1}$ from the previous frame allows the selection of a subset of superpixel nodes $\mathcal{M}_{t-1}^S \subseteq \mathcal{S}_{t-1}$ as seeds for IFT-SC in

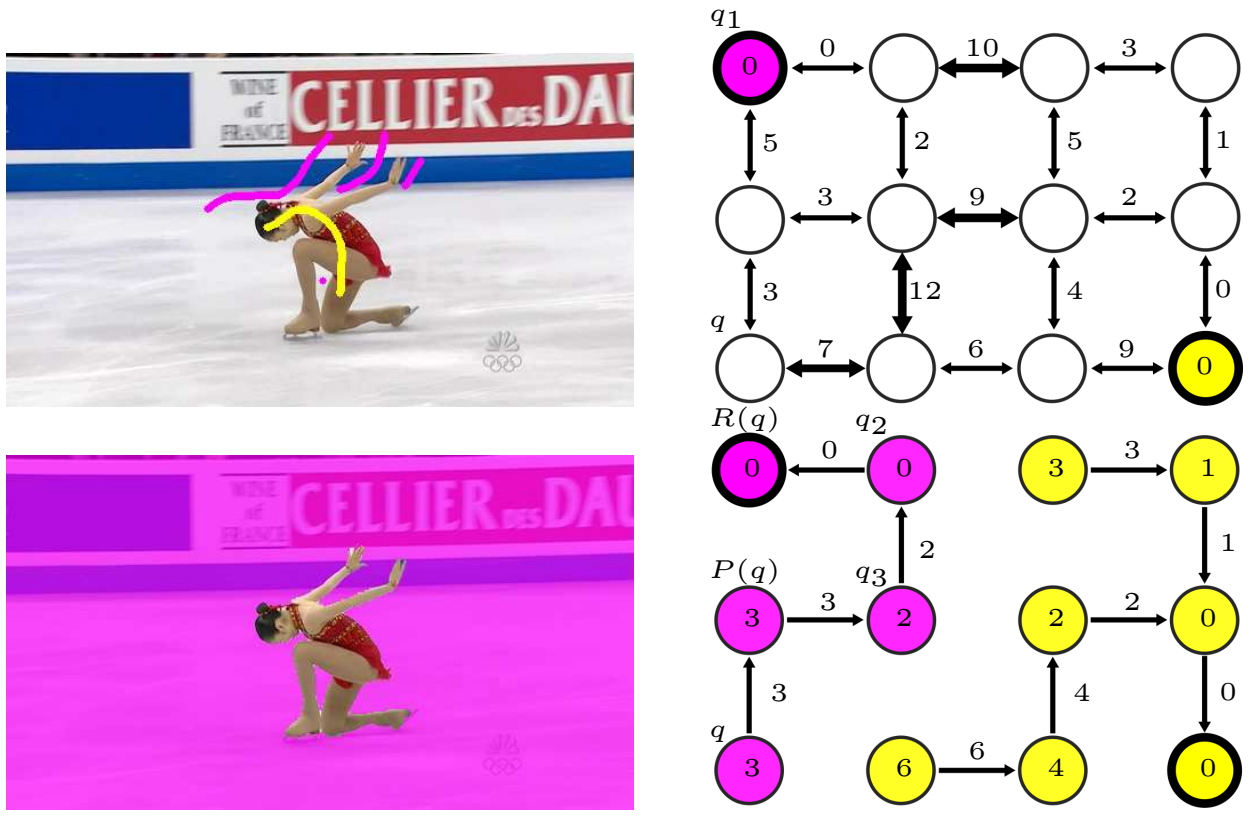

Fig. 4: **Top row**: user-drawn scribbles for interactive image segmentation using IFT-SC, and pixel-graph illustration with weighted arcs between 4-neighbors and two labeled seeds (bold circles). **Bottom row**: segmentation result from the scribbles (left) and optimum-path forest $P$ obtained by IFT-SC from the seed nodes (right). The arcs in this case indicate the predecessor of each node in the forest $P$ (e.g., $P(q_3) = q_2$). The values inside the nodes are the optimum path costs and the node colors indicate the labels propagated from the roots.

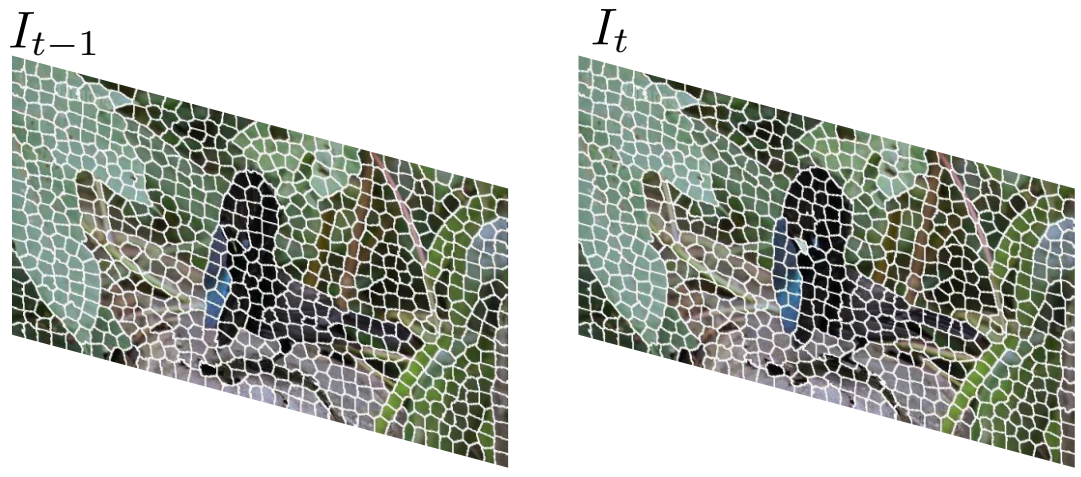

(a) Superpixel boundaries for two consecutive frames.

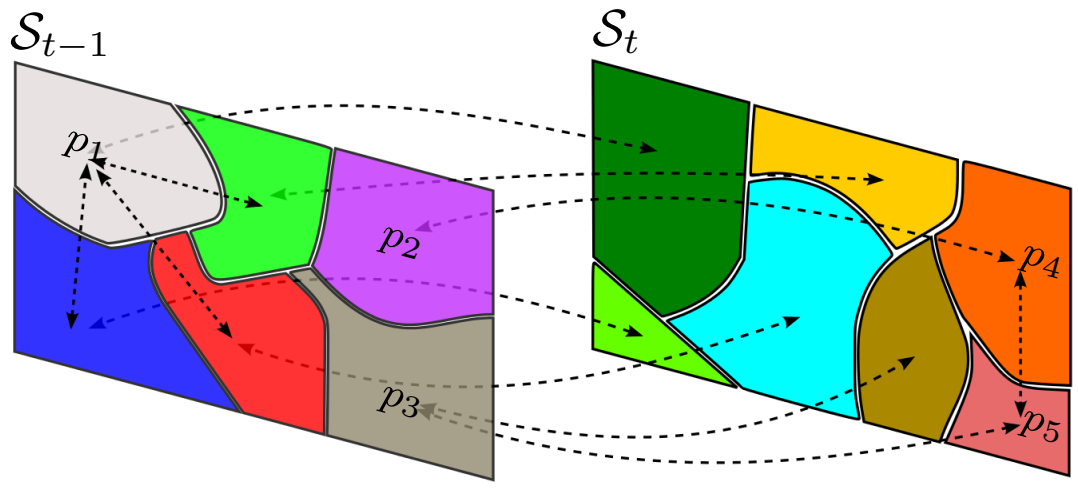

(b) Superpixel-graph example.

Fig. 5: Superpixel-graph creation for two consecutive frames $I_{t-1}$ and $I_t$. We display only a subset of the graph's arcs connecting superpixel nodes from $\mathcal{S}_{t-1}$ and $\mathcal{S}_t$ for clarity.

Eq. 1. FOMTrace constructs seed set $\mathcal{M}_{t-1}^S$ automatically, from the erosion and dilation of $L_{t-1}$ using small radii of $\rho_e = 2$ and $\rho_d = 3$ pixels, respectively. A true superpixel segmentation label $\lambda(p) \in \{0, 1\}$ is given to every node $p \in \mathcal{M}_{t-1}^S$ according to the region that entirely contains the superpixel (foreground or background).[1] Such definition of

[1]As in [17], a superpixel may be selected by pixels from both labels simultaneously. We assign the label with highest number of seed pixels in this case. In the future, we aim to solve this problem hierarchically.

$\mathcal{M}_{t-1}^S$ ensures a dense selection of superpixels that covers most of the foreground and background from $I_{t-1}$ (Figure 6).

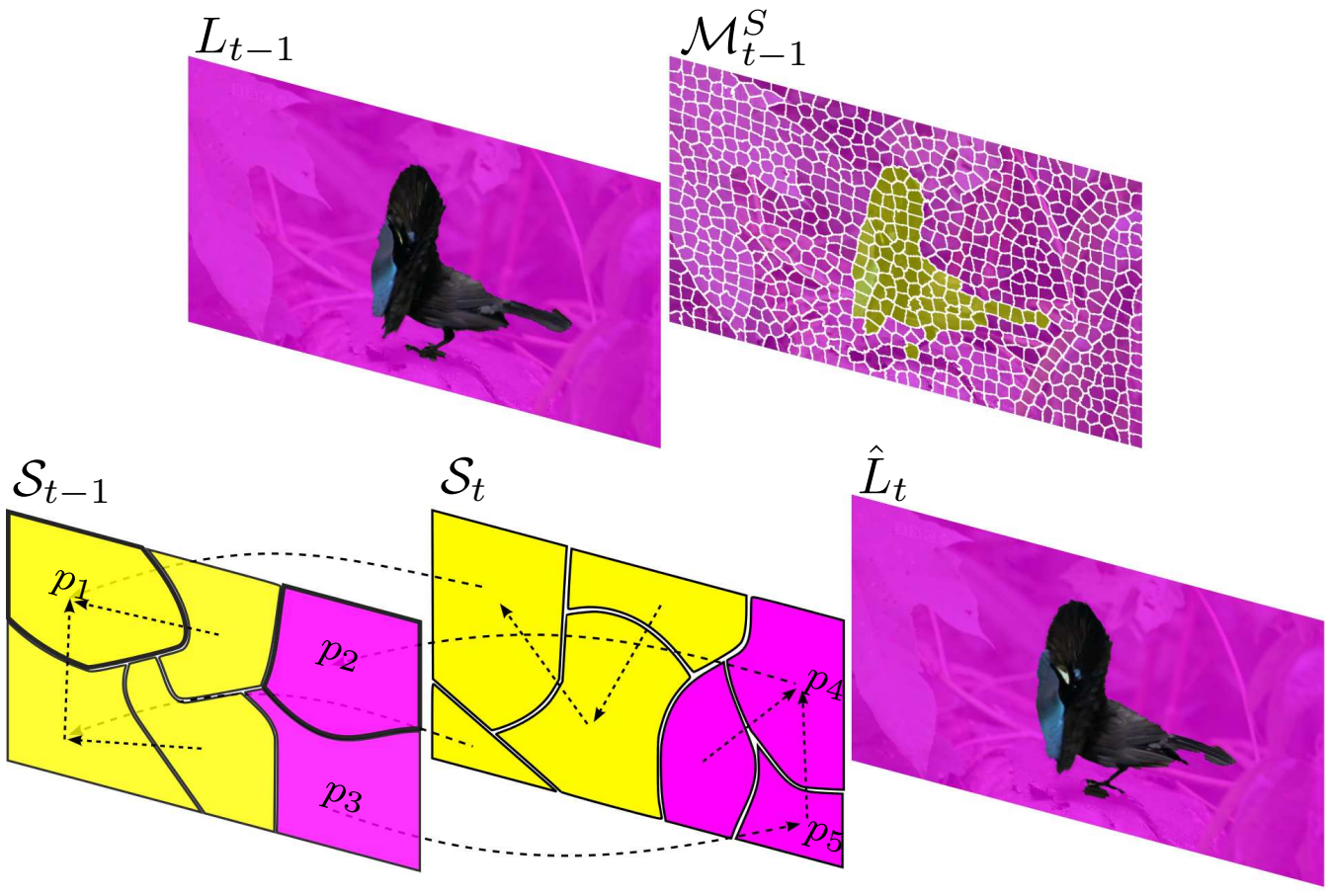

Fig. 6: A segmentation example depicting the previous segmentation result for label $L_{t-1}$, the superpixel seed nodes $\mathcal{M}_{t-1}^S$ selected and labeled according to $L_{t-1}$, the resulting optimum-path forest using IFT-SC on a fictious graph, and the corresponding predicted segmentation label $\hat{L}_t$. The fictious graph depicts two nodes $p_1$ and $p_2$ as seeds from $\mathcal{M}_{t-1}^S$ (bold edges). Note how node $p_3$ was conquered by seed $p_2$ using a path in the optimum-path forest that goes backwards in time through nodes $p_5, p_4 \in \mathcal{S}_t$. (Best viewed in color)

Foreground and background seed regions then compete between themselves in IFT-SC to conquer the remaining superpixels $\mathcal{N} \backslash \mathcal{M}_{t-1}^S$, producing a superpixel label map for all nodes in $\mathcal{N}$, as in the pixel case. Predicted pixel label images $\hat{L}_t, \hat{L}_{t+1}, \ldots, \hat{L}_{n_f-1}$ are readily obtained by assigning to every pixel in $D_{I_t}$ the corresponding superpixel segmentation label, for every frame $t > 0$ (Figure 6). In practice, FOMTrace only requires $\hat{L}_t$ to be computed and refined using model-based segmentation before displaying it to the user. The remaining labels are produced as a side effect and not usually displayed to the user, since they haven't been refined.

**Related works:** The IFT algorithm is agnostic to the graph structure used for segmentation. Hence, we could use virtually *any* existing algorithm for video supervoxel creation [34]–[38] and build a suitable graph out of the result. We take the aforementioned approach because it is easily parallelizable for multi-core CPUs. Moreover, Galasso et al. [39] concluded that graphs of frame superpixels connected via optical flow allow higher boundary accuracy in video segmentation, when compared to techniques that rely on video supervoxels, at the expense of lower temporal consistency. The latter is mitigated by IFT-SC. It is also worth noting that Gangapure et al. [40] have developed a technique for causal video segmentation with similarities to FOMTrace. After automatically segmenting the first video frame into salient regions, they build superpixel frame graphs for the current and next frames for label propagation. This is achieved by propagating superpixel seeds via graph matching, which are then used as markers for watershed on a pixel-graph to segment the next frame. The key differences are that they do not have future frames at their disposal, nor rely on object models for refinement.

### C. Fuzzy Model Computation and its Use for Refinement

*1) Fuzzy Object Modeling:* A Fuzzy Object Model [18]–[22] is a mathematical representation of an object's shape, and possibly texture [20], [41], traditionally composed of a fuzzy image $O_t : D_{I_t} \rightarrow \mathbb{R}$ whose pixels are given grayscale values representing: object (white), background (black), and uncertainty (gray). The uncertainty region encodes expected shape variations, being related to the membership that a pixel has to the foreground. Therefore, this region is a shape prior that constraints the area where the object's real boundary is expected to be when the model is placed at its center in an image under delineation [18], [19].

FOMTrace's training dataset for computing $O_t$ contains two masks: the predicted label image through semi-automatic video segmentation $\hat{L}_t$ and the label image $\widetilde{L}_t$, which corresponds to $L_{t-1}$ propagated from the previous frame $I_{t-1}$ to $I_t$ via dense optical flow [31] (Figures 7b-7c). The previous label image $L_{t-1}$ is the most recent and reliable information regarding the object's silhouette, since in theory it has been accepted as correct by the user. Therefore, it is paramount as a shape constraint to prevent leakings due to color similarities between foreground and background, such as the handlebar delineation error in Figure 7a. We propagate $L_{t-1}$ to $I_t$ by applying the mean optical flow displacement of pixels that belong to a same superpixel $\mathcal{S}_{t-1}^i$. Such propagation places $\widetilde{L}_t$ in the same coordinate space as $\hat{L}_t$, where the object has already been approximately located through video delineation.

We compute the signed Euclidean distance transform [42] (sEDT) from the object mask boundaries of both $\hat{L}_t$ and $\widetilde{L}_t$ resulting in images $\hat{E}_t$ and $\widetilde{E}_t$, respectively (Figures 7d-7e).[2] These images simulate the existence of a larger training dataset, which at the same time better fits the object's dynamic shape observed in both masks. We average $\widetilde{E}_t$ and $\hat{E}_t$ in the following manner to compute a fuzzy image $O_t$:

$$\begin{aligned} O_t(p) = & [\widetilde{E}_t(p).\widetilde{W}_t(p) + \\ & \hat{E}_t(p).(1.0 - \widetilde{W}_t(p))].\delta(\widetilde{L}_t(p) - \hat{L}_t(p)), \end{aligned} \tag{2}$$

where $\widetilde{W}_t$ is an image that weights the importance of the propagated sEDT for computing $O_t$ and $\delta(\cdot)$ is the Dirac delta function. All pixels in $\widetilde{W}_t$ may be set with 0.5 to balance the importance of both distance transforms. See the Appendix for a more refined technique for computing the weight image.

The intuition behind Eq. 2 is that corresponding pixels from $\widetilde{L}_t$ and $\hat{L}_t$ that have different labels necessarily belong to the uncertainty region of $O_t$. In this case, the Dirac delta function will force those pixels to have fuzzy value $O_t(p) = 0.0$.[3] The remaining pixels very likely belong to the real foreground or background, and will be assigned positive or negative values accordingly (recall that we consider *signed* distance transforms). Higher positive values indicate that the pixels are farther inside the foreground region of $O_t$, thus having a higher chance of belonging to the real object. The opposite reasoning is valid for the background.

[2]Our signed Euclidean distance transform assigns non-negative distance values for pixels inside the object mask and negative values outside it.

[3]For multiple objects, Eq. 2 naturally holds since we compute the sEDTs for $\widetilde{L}_t$ and $\hat{L}_t$ from the boundary of each object simultaneously. Hence, the resulting fuzzy image $O_t$ works for multiple labels as well.

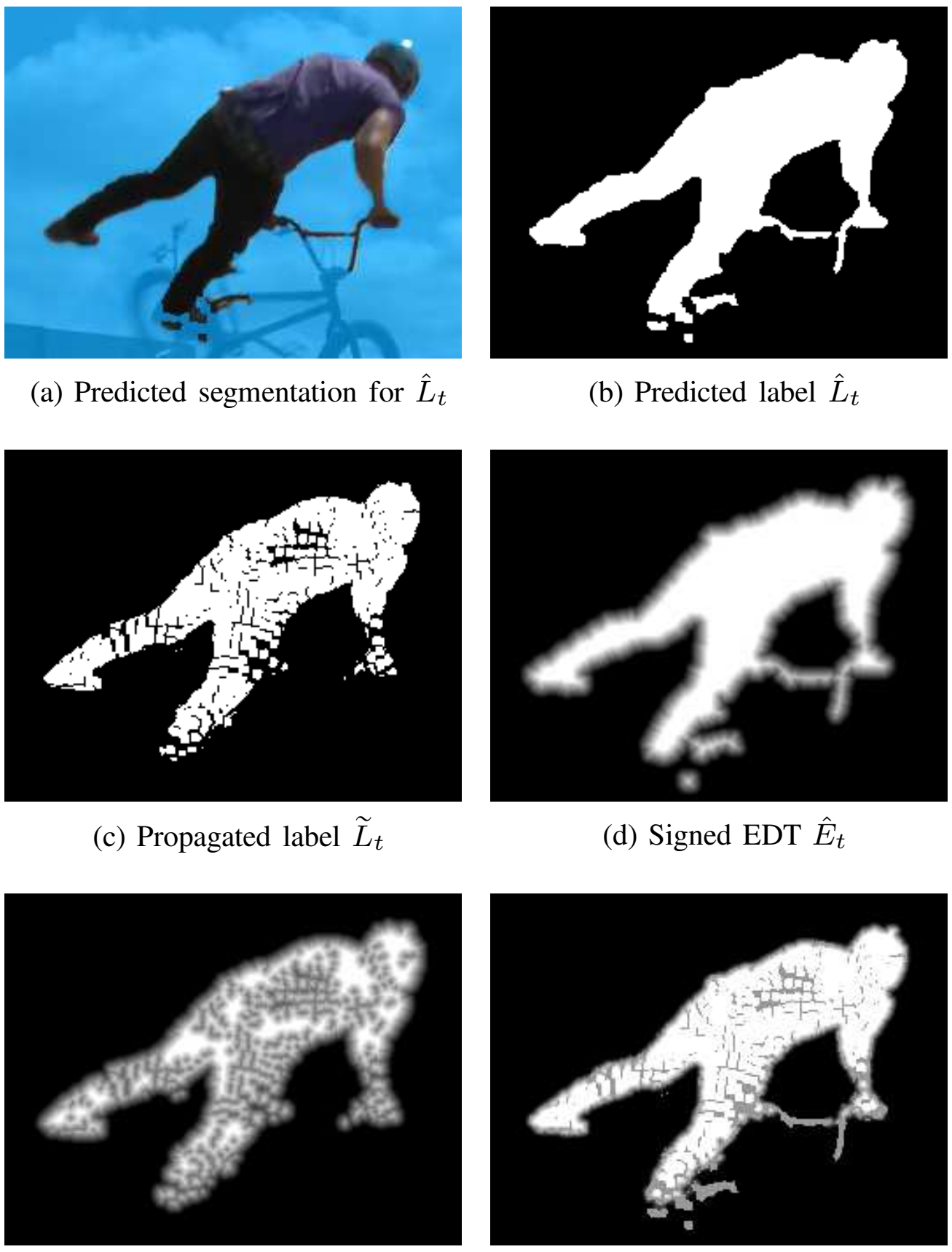

(a) Predicted segmentation for $\hat{L}_t$ (b) Predicted label $\hat{L}_t$

(c) Propagated label $\widetilde{L}_t$ (d) Signed EDT $\hat{E}_t$

(e) Signed EDT $\widetilde{E}_t$ (f) Fuzzy image $O_t$

Fig. 7: (a) Incorrect semi-automatic video segmentation result for label $\hat{L}_t$. (b)-(c) Training set considered for computing the fuzzy image $O_t$. (d)-(e) Signed Euclidean distance transforms (sEDTs) computed from $\hat{L}_t$ and $\widetilde{L}_t$, which are averaged to output fuzzy image $O_t$ in (f).

*2) Model-based Object Refinement:* FOMTrace computes pixel seed set $\mathcal{M}_t^O$ by thresholding $O_t$ using a negative value of $\alpha_b = -2$ for background seeds, and a positive value of $\alpha_f = 3$ for foreground seeds. This operation forces the seeds to be selected beyond a minimum distance to the object's boundary in both masks $\hat{L}_t$ and $\widetilde{L}_t$, thereby increasing the size of the uncertainty region.

Segmentation refinement applies IFT-SC on the same pixel-graph used for interactive image segmentation, this time with seed set $\mathcal{M}_t^O$ (Figure 8a). Note that seed "holes" may occur inside the object and background due to our label propagation via optical flow. They are automatically closed by IFT-SC.

At this point, the refined segmentation label $L'_t$ is finally displayed to the user (Figure 8b), who may correct it interactively (Figure 1). Note in Figures 7a and 8b how model-based image segmentation fixed several leakings from $\hat{L}_t$, although the biker's foot was still lost. The next section (II-D) details how the user may perform further corrections when desired.

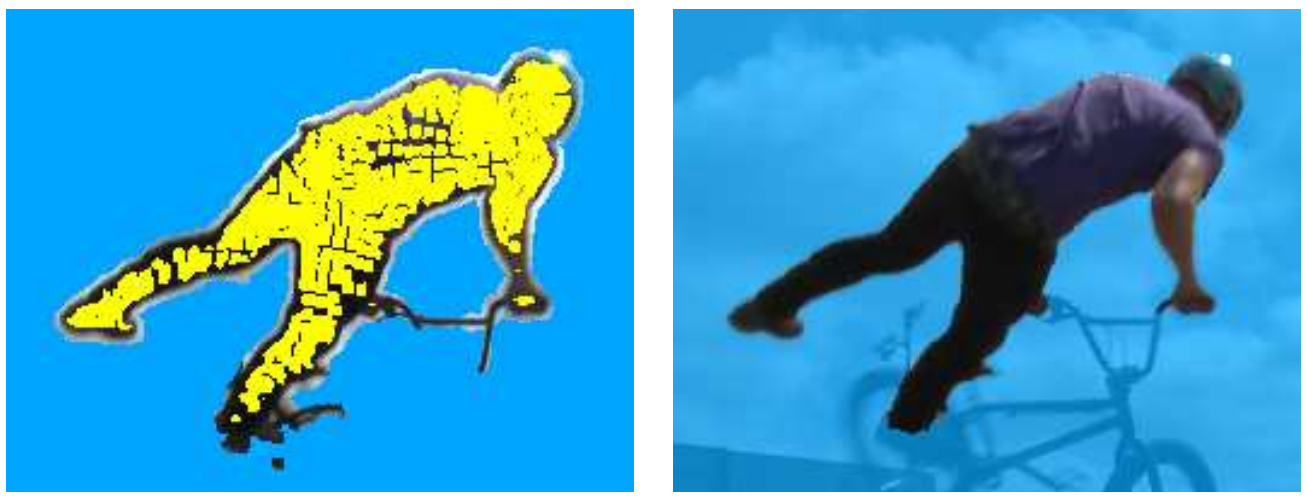

(a) Model seed set $\mathcal{M}_t^O$ (b) Refined segmentation result $L'_t$

Fig. 8: Model-based image segmentation via IFT-SC to refine the semi-automatic video segmentation result $\hat{L}_t$ from Figure 7a. Fuzzy image $O_t$ from Figure 7f provides a set of foreground and background pixel seeds $\mathcal{M}_t^O$ that is used by IFT-SC to compute a new refined segmentation mask $L'_t$.

**Related works:** In medical image segmentation, a FOM is used to locate and delineate a given object of interest in a new target image automatically [18]–[22]. Our case is similar except that we have already approximately located the object through delineation and need only to fix possible leakings using $O_t$. Our fuzzy object modeling approach can be seen as an improvement over techniques that create a similar fuzzy shape image, although considering only the combination of shapes from past frames [14], as in medical imaging, or only the propagated label [10]–[12]. The approach in [10], [11] represents the state-of-the-art in interactive video segmentation and is part of the commercial software Adobe After Effects. We have compared FOMTrace with both techniques [14] and [10], [11] in our experiments. We further note that approaches such as *Watershed from popagated markers* [8] also tend to use shape constraints to prevent leakings.

### *D. Interactive Correction in FOMTrace*

The model-based object refinement produces an optimum-path forest $P_t$ from the image $I_t$, at time $t > 0$, which is rooted at the interior and exterior pixels of the fuzzy object model $O_t$ (Figure 8a). The corresponding segmentation results in the label map $L'_t$ (Figure 8b). In the case of errors, the user can add seeds or remove trees to correct segmentation, without starting over the process, by using the differential version of IFT-SC [16] and other IFT-derived methods [7]. However, corrections are more effective when the root set is considerably smaller than those from $O_t$, so the user can decide to add seeds or remove trees more easily. For that purpose, we use the method proposed in [15] to convert $L'_t$ into an optimum-path forest with minimum number of roots (Figure 9).

**Related works:** Several works consider the propagation of segmentation directly from the user's scribbles, without requiring a previous segmentation mask as input [9], [13], [44]–[46]. Some of these even treat the video directly as a 3D image that can be and manipulated viewed and manipulated as such [45], [46]. We believe that frame-by-frame video segmentation provides more user control, while allowing object shape models to be created and used to prevent leakings. Nevertheless, FOMTrace is flexible enough in that the user

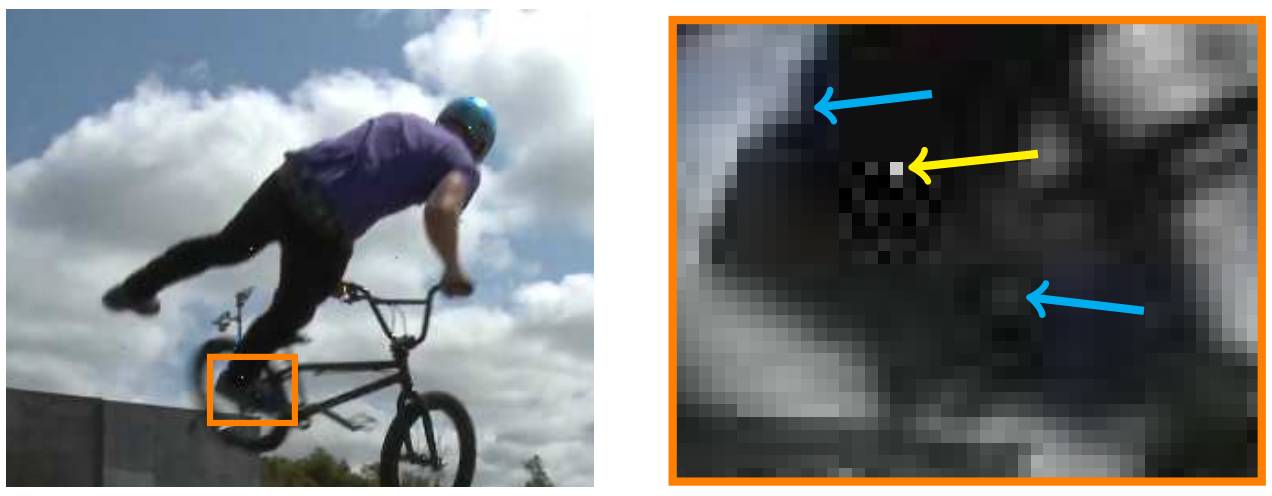

(a) Resuming seed set for DIFT-SC computed from $L'_{t=5}$ using the method in [15], [43]. The arrows point out automatically selected foreground and background seeds for easier reference.

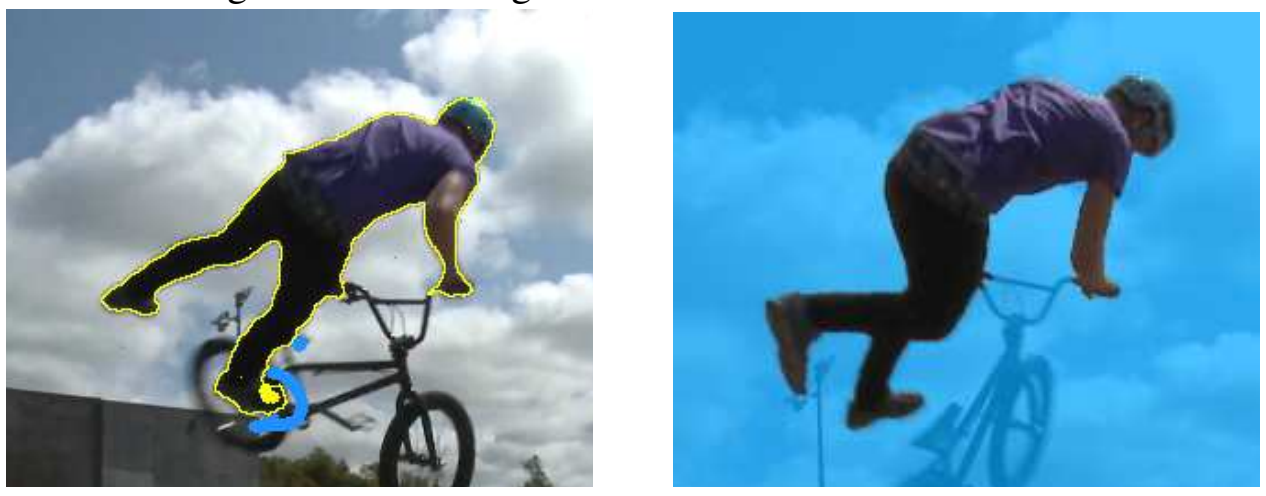

(b) Final segmentation. $L_5$ after user corrections.

(c) FOMTrace's segmentation label $L_{25}$ after correction.

Fig. 9: Interactive correction of FOMTrace's segmentation for the frame from Figure 8. (a) Instead of having the user-drawn scribbles compete with the densely populated FOM seed set $\mathcal{M}^O_{t=5}$ (Figure 8a), we reconstruct an optimum-path forest whose labeling is equivalent to $L'_5$ but that requires a reduced seed set [15], [43]. (b) The user selected scribbles compete against the automatically added seeds. (c) FOMTrace's result for frame $t = 25$, after correcting frame $t = 5$.

may choose not to use our fuzzy object model, performing segmentation directly on the superpixel-graph by selecting superpixel nodes with scribbles.

## III. Experimental Evaluation

We have evaluated FOMTrace in two standard datasets for testing video object tracking and segmentation methods, namely, the SegTrackv2 [47] and IFTrace [48] datasets. Both datasets provide dense pixelwise ground truth annotation for all videos, which range in length from 21 to 753 frames. Li et al. [47] designed SegTrackv2 to assess methods for automatic and semi-automatic video object segmentation. It contains videos charecterized by: fast object motion, motion blur, appearance change, complex deformation, partial occlusions, slow motion, and multiple interacting objects. Minetto et al. [48] designed the IFTrace dataset to evaluate techniques for object tracking. The main challenges aim to stress those techniques in common situations that occur during tracking and include: partial and total occlusions, deformable objects, illumination changes, multiple objects, and low contrast between the foreground and the background. We have compared FOMTrace with the automatic video segmentation, named Segment Pool Tracking + Composite Statistical Inference [47] (SPT+CSI), semi-automatic video delineation methods, IFTrace [48] and the method of Gallego and Bertolino [14], and Rotobrush, the interactive segmentation tool from Adobe After Effects, which uses the Video SnapCut method [10], [11].

In the interactive/semi-automatic case, the standard procedure for computing accuracy in those datasets is to initialize the method using the annotated label in the first frame, and then to run it uninterruptedly for the remainder of the video. This is a simple variant of FOMTrace, when the user is not required to verifiy the result. It should be clear that under the user's supervision FOMTrace achieves much better results.

The SegTrackv2 standard evaluation protocol considers the average *intersecion over union score* ($\overline{IoU}$) to measure accuracy for each video [47], being calculated for a single frame as $IoU = \frac{TP}{TP+FP+FN}$, where $TP, TN, FP, FN$ are the true/false positive/negative scores. In the IFTrace dataset the procedure is similar, but the $F1$ score is the standard metric instead [48]: $F1 = \frac{2\times P\times R}{P+R}$, where $P$ and $R$ are the *precision* and *recall* rates, respectively.

We present qualitative results that demonstrate the evolution of segmentation across time for some of the semi-automatic/interactive methods. We also include interactive experiments with FOMTrace and FOMTrace$^{\mathrm{W}}$ following the benchmark metrics proposed in [8]. Those metrics include the amount of user interaction and time spent in the segmentation process for each frame, which are difficult to measure in commercial softwares such as Adobe After Effects. Hence, we were unable to obtain reliable measures for Rotobrush and therefore do not compare with it. See our supplementary material for video results obtained interactively with FOMTrace, on sequences that present multiple challenges.

### A. *Parameter settings*

We have used the parameter-free version of SLIC named SLICO [33] in our implementation. We compute SLICO superpixels using a regular image grid initilization with steps between 5 pixels and about 1% of the frame's width/height, depending on the object and frame dimensions. This guarantees small superpixels that properly adhere to the object's boundary in most cases (we used 5 pixels steps in our experiments).

Another important parameter of our method is the threshold $\gamma$ used in Eq. 3 to compute fuzzy image $O^W_t$. The FOMTrace variant in the Appendix uses a more restrictive fuzzy image $O^W_t$ as shape constraint, which ensures a stronger participation of $\widetilde{L_t}$ to prevent leakings in touching regions of foreground and background with high color similarity. Parameter $\gamma$ controls the strictness of this participation and should be set differently by the user throughout each video sequence. For the automatic comparison with the state-of-the-art, we have empirically validated that $\gamma = 0.6$ helps in those situations. We define FOMTrace as the default configuration of our method using only Eq. 2 with a balanced weight image, and FOMTrace$^{\mathrm{W}}$ as the variant weighted by $\widetilde{W_t}$ in Eq. 3 with $\gamma = 0.6$.

### B. *Diagnostic experiments*

Since we propose FOMTrace as a method that uses fuzzy objects to fix volumetric video segmentation at each frame, we first performed some diagnostic experiments comparing FOMTrace and FOMTrace$^{\mathrm{W}}$ with Superpixel IFT-SC. Superpixel IFT-SC corresponds to running FOMTrace's automatic video segmentation step from the first video frame only, selecting the seeds from $L_0$ as in Section II-B to determine the labels for the rest of the video $\hat{L}_1, \hat{L}_2, \ldots, \hat{L}_{n_f-1}$ in a single shot, *without* model-based refinement. Table I presents these results.[4]

TABLE I: Diagnostic experiments comparing the usage of FOMs to correct volumetric superpixel delineation in the SegTrackv2 dataset [47]. Best scores are in bold.

| Video | Superpixel IFT-SC | FOMTrace | FOMTrace$^W$ |
|---|---|---|---|
| bird_of_paradise | 87.1 | **89.9** | 24.0 |
| birdfall | **60.2** | 58.5 | 34.8 |
| bmx 1 | 84.0 | **94.2** | 69.2 |
| bmx 2 | **5.2** | 4.6 | **5.2** |
| cheetah 1 | 12.1 | 9.1 | **14.9** |
| cheetah 2 | **24.7** | 21.0 | 21.4 |
| drift 1 | 67.3 | **78.3** | 75.4 |
| drift 2 | 17.2 | **51.3** | 30.7 |
| frog | **53.8** | 40.6 | 47.1 |
| girl | 76.1 | **78.0** | 56.0 |
| hummingbird 1 | 18.2 | 18.0 | **21.4** |
| hummingbird 2 | **44.5** | 42.3 | 30.2 |
| monkey | 66.9 | **85.5** | 29.4 |
| monkeydog 1 | **81.5** | 80.1 | 64.5 |
| monkeydog 2 | 67.2 | 66.0 | **78.4** |
| parachute | 78.6 | **93.6** | 84.0 |
| penguin 1 | 91.9 | 95.4 | **96.2** |
| penguin 2 | 85.3 | **94.0** | 93.1 |
| penguin 3 | 85.8 | **89.3** | 89.1 |
| penguin 4 | 75.5 | 87.1 | **88.1** |
| penguin 5 | 73.4 | 83.8 | **86.7** |
| penguin 6 | 84.8 | 88.8 | **89.9** |
| soldier | 65.3 | **79.0** | 76.6 |
| worm | 6.4 | **80.2** | 12.3 |

Table I shows that segmentation accuracy improves for most video sequences when using FOMs to correct frame delineation. In 18 out of 24 videos, the shape restriction of FOMs, using either FOMTrace or FOMTrace$^W$, prevented leakings. In contrast, Superpixel IFT-SC works better in the absence of leakings and for highly deformable fast moving objects, as in videos `frog`, `hummingbird 2`, and `monkeydog 1`.

### C. Quantitative comparison with the state-of-the-art

We performed direct and indirect comparisons with the state-of-the-art. Specifically, SPT+CSI [47] is the baseline for the SegTrackv2 dataset, and the method of Gallego and Bertolino [14] was validated in the same dataset. Therefore, we make an indirect comparison with those methods by reproducing here the results reported in their papers for easier reference. The comparison with IFTrace [48] used our C/C++ implementation. Finally, we used the latest Adobe After Effect CC 2014's Rotobrush tool to perform our experiments.

We were unable to load the ground truth mask as input for Rotobrush. Therefore, we separate the evaluation into two parts for the SegTrackv2 dataset. Table II compares SPT+CSI and Gallego [14] with the result of running FOMTrace and IFTrace from the first frame's ground truth mask. For clarity, we omit FOMTrace$^W$ in this evaluation and refer the reader to Table I for the corresponding values, specially for video `monkeydog 2`. This is a difficult example with fast camera motion and high color overlap between foreground and background, in which FOMTrace$^W$ outperformed all other methods.

[4]We consider best scores within 0.1 points to be equivalent.

In 11 out of the 24 sequences FOMTrace achieved the best score, while being close to the state-of-the-art for other cases (e.g., `birdfall`, `soldier`, `worm`).

TABLE II: Comparing FOMTrace with methods for unsupervised and semi-automatic segmentation in the SegTrackv2 dataset. The values refer to the mean *intersection over union* scores for each video sequence and method. Results for SPT+CSI [47] and Gallego [14] are reported from the original works for reference.

| Video | SPT+CSI [47] | Gallego [14] | IFTrace [48] | FOMTrace |
|---|---|---|---|---|
| bird_of_paradise | 94.0 | **95.4** | 14.5 | 89.9 |
| birdfall | **62.5** | 59.6 | 15.8 | 58.5 |
| bmx 1 | 85.4 | 86.6 | 43.2 | **94.2** |
| bmx 2 | **24.9** | 0.0 | 2.8 | 4.6 |
| cheetah 1 | **37.3** | 30.1 | 10.8 | 9.1 |
| cheetah 2 | **40.9** | 20.5 | 14.0 | 21.0 |
| drift 1 | 74.8 | **85.4** | 59.9 | 78.3 |
| drift 2 | 60.2 | **72.4** | 70.5 | 51.3 |
| frog | 72.3 | **74.5** | 53.4 | 40.6 |
| girl | **89.2** | 87.9 | 72.2 | 78.0 |
| hummingbird 1 | **54.4** | 26.2 | 30.7 | 18.0 |
| hummingbird 2 | **72.3** | 59.2 | 23.4 | 42.3 |
| monkey | 84.8 | 75.1 | 23.4 | **85.5** |
| monkeydog 1 | 71.3 | **80.2** | 11.5 | **80.1** |
| monkeydog 2 | 18.9 | 48.3 | 57.3 | **66.0** |
| parachute | 93.4 | **93.6** | 92.4 | **93.6** |
| penguin 1 | 51.5 | **95.4** | 60.8 | **95.4** |
| penguin 2 | 76.5 | 89.4 | 53.7 | **94.0** |
| penguin 3 | 75.2 | 81.1 | 54.1 | **89.3** |
| penguin 4 | 57.8 | 80.6 | 50.3 | **87.1** |
| penguin 5 | 66.7 | 76.3 | 73.2 | **83.8** |
| penguin 6 | 50.2 | 78.0 | 49.7 | **88.8** |
| soldier | **83.8** | 76.7 | 69.4 | 79.0 |
| worm | **82.8** | 53.3 | 31.1 | 80.2 |

We present the second part of the experiments in the SegTrackv2 in Table III. To compare with Rotobrush in a fair way, guaranteeing pixel-level accuracy, we interactively initialized Rotobrush in the first frame and let it propagate segmentation for the remainder of the video, for every sequence in SegTrackv2. We ensured that the interactive initialization was as close as possible to the original ground truth. We then applied the corresponding first mask as input for IFTrace, FOMTrace, and FOMTrace$^W$. As one may note by comparing Tables I, II, and III, the differences are minute but may determine the best score for a certain video sequence.

From Table III we see that FOMTrace and FOMTrace$^W$ achieved the highest score in 14 out of 24 video sequences. At this point, one may argue that FOMTrace$^W$ is a fine tuned version of FOMTrace, and that Rotobrush could obtain equivalent improvement via parameter setting. Although this is a valid assumption, if we disconsider FOMTrace$^W$'s results FOMTrace still achieves higher accuracy in 12 cases, versus 10 for Rotobrush and 4 for IFTrace. Moreover, we note that using FOMTrace$^W$ with $\gamma = 0.6$ actually decreases performance in some cases, confirming our intuition that FOMTrace should be used by default, letting the user determine the best time to use FOMTrace$^W$. Such situations often occur when the foreground has strong color overlap with the background, but the optical flow estimation is reliable enough to allow the simple propagation of the previous mask to the current frame. For example, sequences `penguin 1-6` refer to distinct animals in the same video that are walking side-by-side towards the camera. Hence, there is overlap between their colors but the optical flow allows the reliable propagation of the previous mask to the current frame for improved label correction.

TABLE III: Comparing FOMTrace with Adobe After Effects Rotobrush tool [10], [11] for interactive video segmentation and IFTrace [48] in the SegTrackv2 dataset.

| Video | Rotobrush [11] | IFTrace [48] | FOMTrace | FOMTrace$^W$ |
|---|---|---|---|---|
| bird_of_paradise | 80.8 | 32.0 | **89.8** | 22.8 |
| birdfall | 2.6 | 18.3 | **58.2** | 32.2 |
| bmx 1 | 90.4 | 46.8 | **94.1** | 72.0 |
| bmx 2 | 2.6 | 2.6 | **6.1** | 5.0 |
| cheetah 1 | **16.0** | 11.4 | 8.9 | 13.9 |
| cheetah 2 | **25.8** | 15.6 | 20.8 | 20.6 |
| drift 1 | **79.8** | 69.5 | 78.3 | 75.4 |
| drift 2 | 50.4 | **74.4** | 51.2 | 34.6 |
| frog | 45.6 | **55.3** | 40.6 | 47.1 |
| girl | 63.1 | 73.6 | **79.0** | 59.5 |
| hummingbird 1 | 21.8 | **28.2** | 17.7 | 20.9 |
| hummingbird 2 | **43.4** | 36.8 | 42.2 | 30.2 |
| monkey | 83.8 | 22.3 | **85.4** | 33.7 |
| monkeydog 1 | 76.0 | 48.1 | **80.0** | 64.5 |
| monkeydog 2 | 71.5 | 14.7 | 65.6 | **76.9** |
| parachute | **94.2** | **94.1** | 93.6 | 84.0 |
| penguin 1 | **96.5** | 57.7 | 95.4 | 96.1 |
| penguin 2 | 93.3 | 49.8 | **94.0** | 93.1 |
| penguin 3 | **89.4** | 50.8 | **89.3** | 89.0 |
| penguin 4 | 85.7 | 51.2 | 87.0 | **87.8** |
| penguin 5 | 84.5 | 72.8 | 83.8 | **86.6** |
| penguin 6 | 87.9 | 45.8 | 89.0 | **89.9** |
| soldier | 70.6 | 72.7 | **79.0** | 76.0 |
| worm | **84.3** | 41.3 | 80.1 | 9.2 |

Finally, Table IV presents the mean $F1$ scores for Rotobrush, IFTrace, FOMTrace, and FOMTrace$^W$ in the IFTrace dataset [48]. As previously stated, this dataset was originally proposed to evaluate IFTrace, which is a semi-automatic video object segmentation method designed for object tracking. Since both Rotobrush and FOMTrace are *interactive* video segmentation techniques, object tracking challenges such as total occlusions are not the primary concern because the user is in control. Nevertheless, it is interesting to observe the methods' behavior under those conditions. In contrast to IFTrace's results in the SegTrackv2 dataset, in which it stood out only 4 times out of 24 (Table III), IFTrace achieves the highest score for 7 videos in its dataset, followed by 4 wins for both Rotobrush and FOMTrace/FOMTrace$^W$.

TABLE IV: FOMTrace versus Rotobrush in IFTrace in the IFTrace dataset [48]. The values for each method refer to the mean $F1$ scores in every video sequence.

| Video | Rotobrush [11] | IFTrace [48] | FOMTrace | FOMTrace$^W$ |
|---|---|---|---|---|
| v01 | 14.3 | **53.1** | 15.0 | 15.1 |
| v02 | 88.3 | **92.7** | 89.8 | 74.8 |
| v03 | 9.4 | **85.4** | 78.8 | 79.8 |
| v04 | 3.4 | 0.7 | **80.3** | **80.3** |
| v05 | 92.7 | **97.8** | **97.7** | **97.7** |
| v06 | 20.7 | **95.4** | 48.4 | 48.4 |
| v07 | 1.7 | **38.1** | 2.4 | 2.7 |
| v08 | **93.6** | 86.2 | 64.8 | 64.9 |
| v09 | **93.5** | 73.9 | 86.8 | 65.5 |
| v10 | 96.3 | 90.7 | **97.3** | 96.7 |
| v11 | **88.5** | 85.4 | 84.9 | 80.8 |
| v12 | **88.4** | 76.9 | 76.6 | 77.7 |
| v13 | 72.9 | 69.6 | 83.3 | **87.0** |
| v14 | 14.6 | **88.1** | 67.4 | 79.8 |

In particular, IFTrace did well in videos v01, v03, v07, and v14. Video v01 depicts a car chase and is characterized by two total occlusions, one between frames $t = 87, .., 90$, when the car passes under a traffic sign, and another between $t = 408, .., 442$, when it speeds under a bridge. IFTrace was designed with a mechanism for retrieving the object after total occlusions, thereby surpassing FOMTrace and Rotobrush in v01. Similarly, IFTrace does better in the other videos because it can handle small objects (v03, v06, v07), low contrast frames (v06, v07), and drastic illumination changes (v14), which are typical issues in object tracking. The user may easily correct segmentation in those cases with FOMTrace, since our goal is to minimize his/her effort for obtaining pixel accurate delineation mainly for fast moving deformable objects and when the foreground and background colors overlap.

Moreover, since we use a fuzzy object model, we can detect when the object goes missing in total occlusions and easily apply the automatic search procedure from [20] to retrieve the object, for example. Also, illumination changes and low contrast can be dealt with by improving the arc weights in our superpixel-graph, which at the moment only considers the mean superpixel color.

### D. Qualitative comparison with the state-of-the-art

Figure 10 depicts some qualitative examples for all four methods we had access to, from the results in Tables III and IV (again, without user correction). The bmx 1 video clip, shown first, presents a fast moving object composed of a person doing maneuvers on a bike. FOMTrace was designed for these situations since the automatic video segmentation step aims to capture fast changes in the object's silhouette. In contrast, FOMTrace$^W$ is more conservative because it relies mostly in the optical flow propagation, when using $\gamma = 0.6$, leading to segmentation errors.

In video clip v13, from the IFTrace dataset, FOMTrace$^W$'s shape restriction due to weighted label propagation helped preserve the person's shape during its partial occlusion by the car, which did not occur for FOMTrace. Hence, videos bmx 1 and v13 demonstrate opposite situations that stress the complementary properties of FOMTrace and FOMTrace$^W$.

In both video sequences of Figure 10, FOMTrace and FOMTrace$^W$ were visually competitive or better than Rotobrush and IFTrace. Figure 11 depicts a failure case for video segmentation using both FOMTrace and FOMTrace$^W$.

### E. Interactive experiments using FOMTrace

Figure 12 depicts graphs with metrics derived from Flores et al. [8] that were devised to measure the accuracy of interactive video segmentation methods and the amount of user effort and time spent in the process, for two videos from the SegTrackv2 dataset. The graphs in the first row present the $IoU$ score for each frame after user corrections. The graphs in the second and third row present the required amount of user interaction and user time for each frame, respectively. The amount of user interaction is given by the number of markers that were added to correct segmentation. We do not count the number of times that the user deleted markers, as opposed to [8], because that kind of interaction mostly serves for the user to correct his/her own mistakes that were made when drawing the scribbles. The amount of user time considers the entire time spent in correcting the frame with our interactive image segmentation tools [7], [15], [43], [49] (up to two minutes per frame in general). We initialized the method with the ground truth segmentation mask as in the automatic experiments.

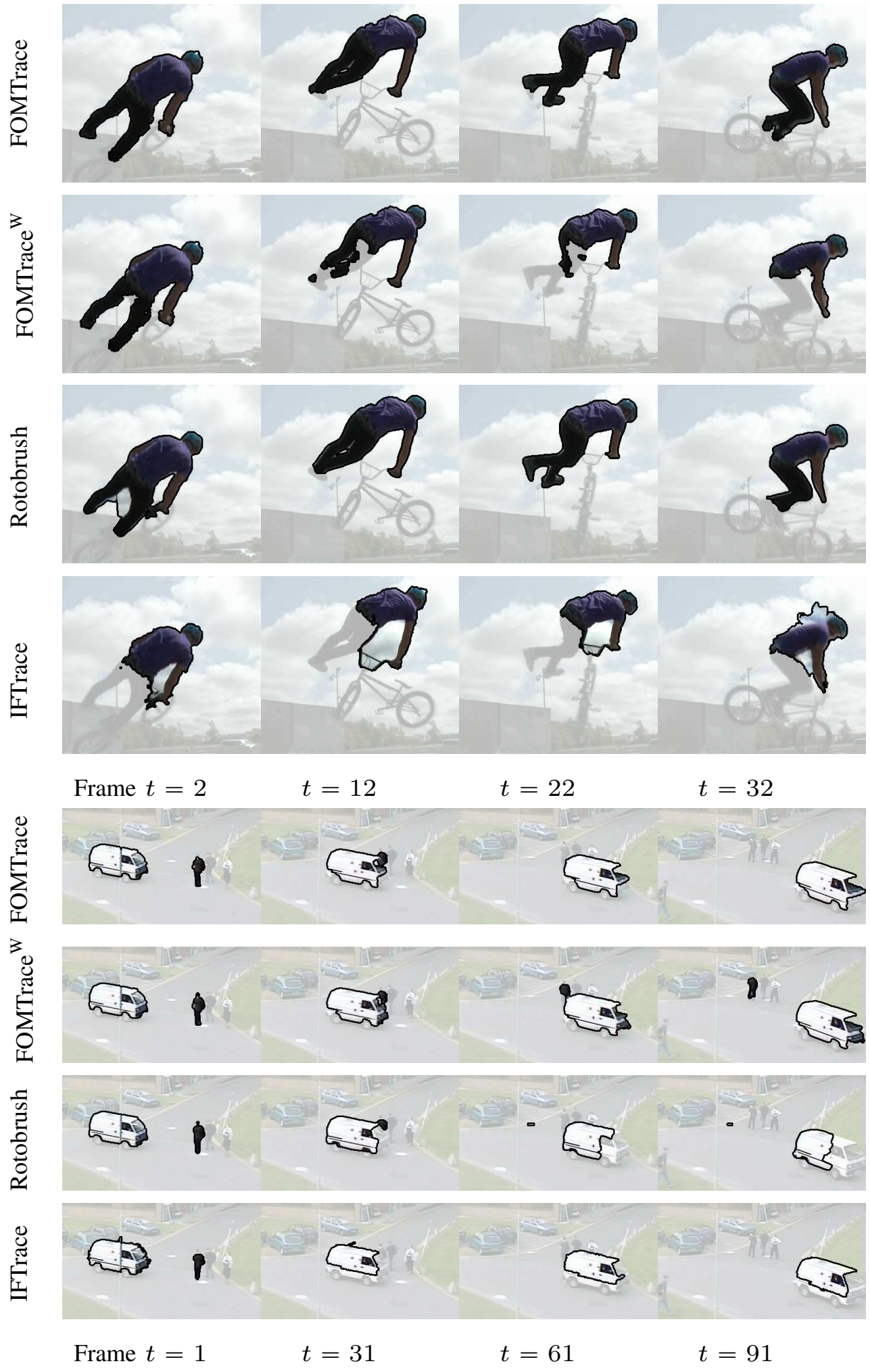

Fig. 10: Qualitative comparison among FOMTrace, FOMTrace$^W$, Rotobrush, and IFTrace in videos `bmx 1`, from SegTrackv2, and `v13`, from the IFTrace dataset.

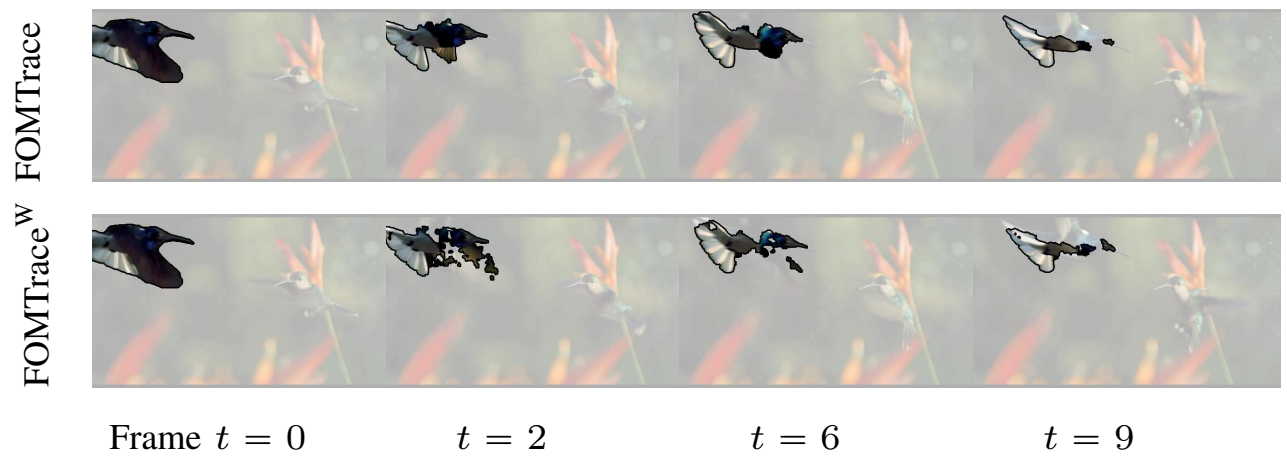

Fig. 11: Failure case for FOMTrace and FOMTrace$^W$ in video `hummingbird 2` from SegTrackv2. Our methods failed mostly due to large fast motion that confused optical flow estimation (on the hummingbird's wings).

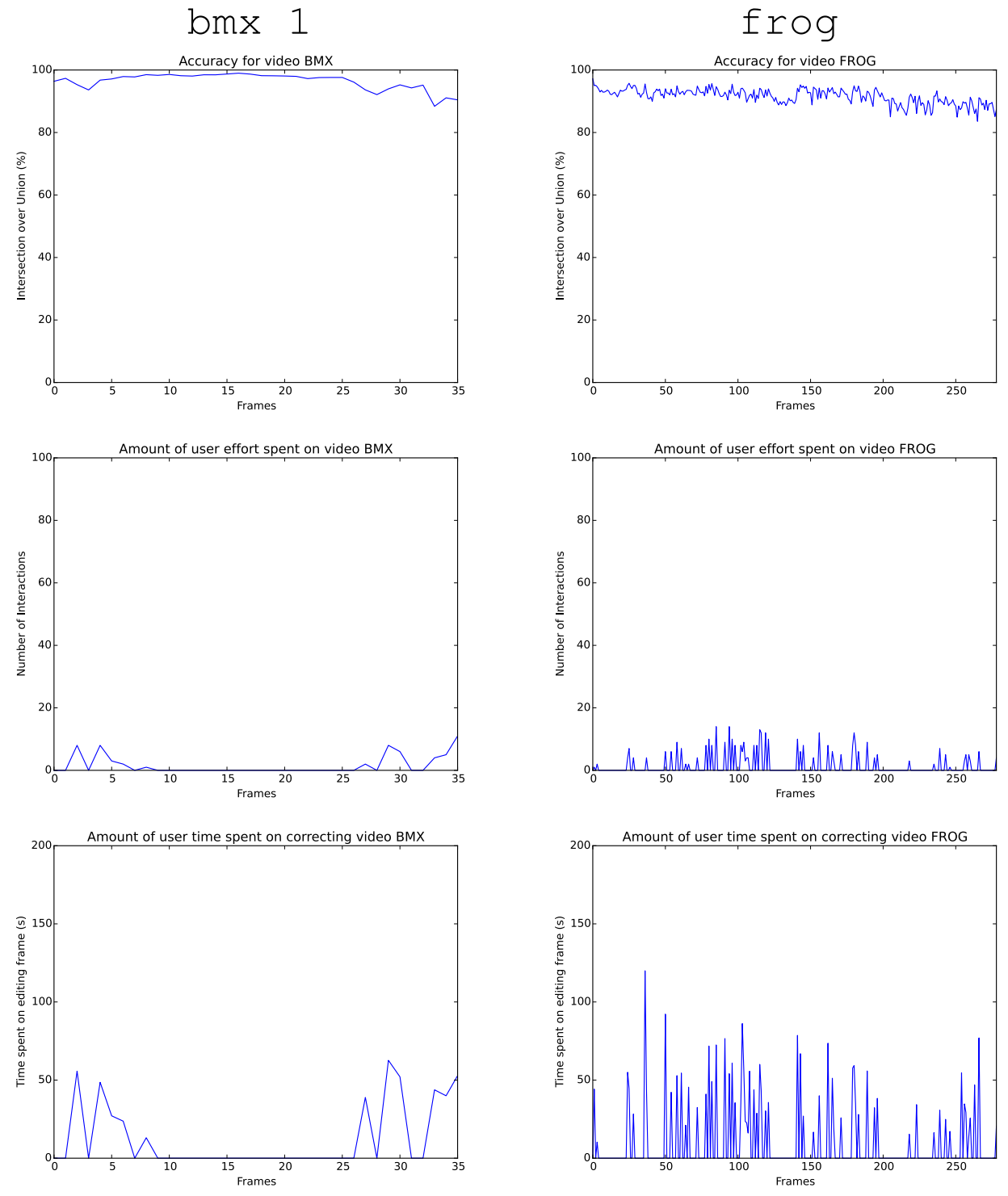

Fig. 12: Interactive results obtained with FOMTrace and FOMTrace$^W$ for two images from the SegTrackv2 dataset: `bmx 1` and `frog`. The graphs in the first row depict the intersection over union score for each frame, the ones in the second row depict the amount of user interaction (i.e., the number of added markers) required for correcting the automatic result, and the ones in the latter row depict the amount of time it took to make the corrections.

From Figure 12, we see that with little interaction the user was able to increase the segmentation accuracy (less than 10 markers per frame in general). In the `bmx 1` video, the mean accuracy was of $96.9\%$ after correcting $36.1\%$ of the 36 frames interactively. The resulting accuracy represents an increase of $2.7\%$ with respect to what FOMTrace achieved without user corrections in Table II. As can be seen from the accuracy graph, the result deteriorated towards the end of the video, which decreased the mean score. This is because the user decided that the ground truth for the latter frames was wrong, since the object presented heavy blur due to fast motion. Hence, the accuracy may be much higher. The second video in our experiment presents a much more dramatic increase in mean accuracy. It went from $40.6\%$ in Table II to $91.6\%$ after user corrections. In this case, FOMTrace$^W$ was intensively used in conjunction with FOMTrace, since there was significant color overlap between foreground and background. The user made corrections only to $23.3\%$ of the 279 video frames to increase the accuracy by $51\%$. Besides intense color overlap, the object (a leaping frog, Figure 13) presented thin structures that were difficult to segment. This hard example demonstrates the power of our framework.

Figure 13 depicts two qualitative results obtained interactively with FOMTrace. The sequences present challenges such as high color overlap between foreground and background and the simultaneous segmentation of multiple objects. See our supplementary material for more video results.

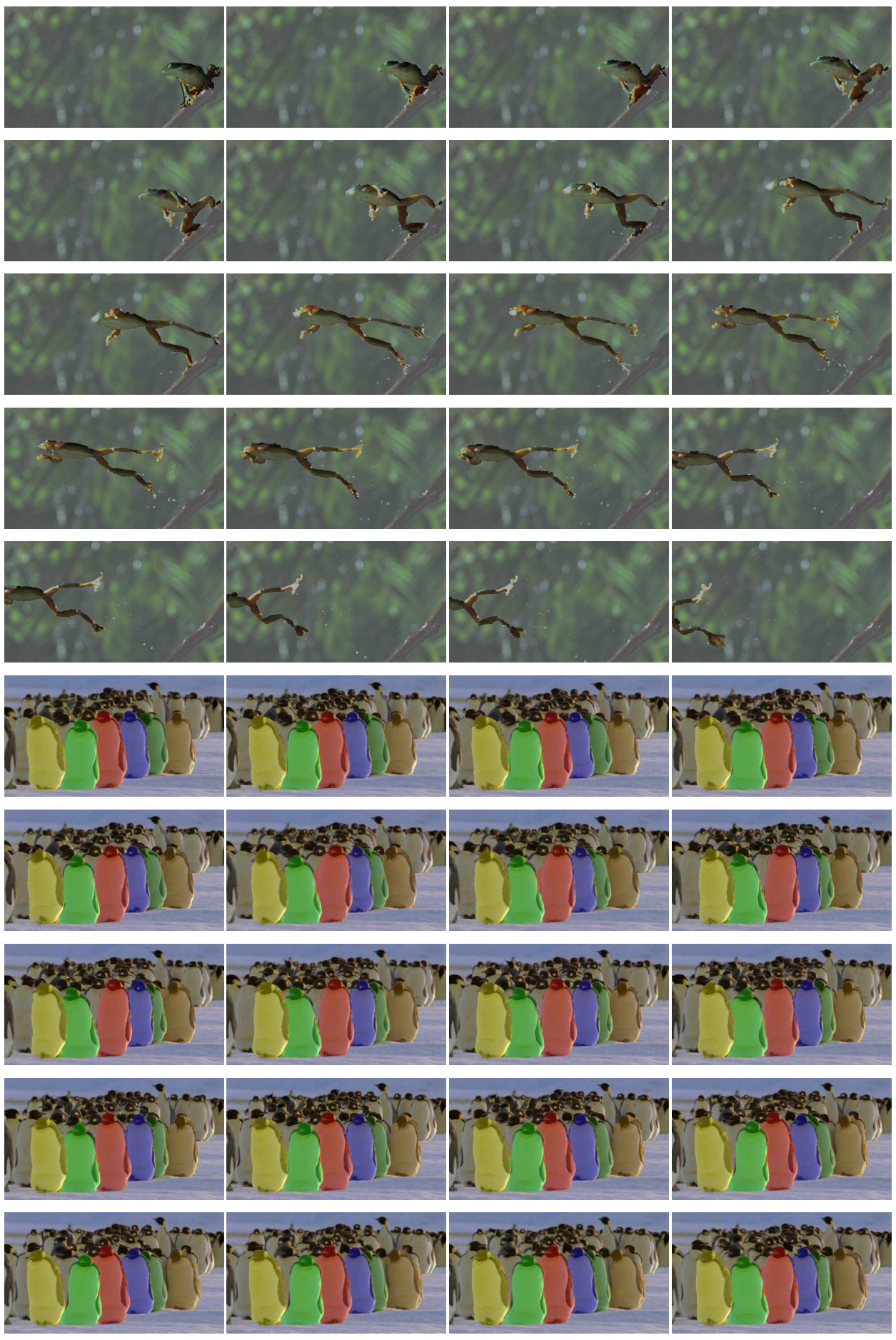

Fig. 13: Interactive FOMTrace results on videos `frog` (top) and `penguin` (bottom). The object in the former presents complex deformation and similar colors to the background. The multiple objects in `penguin` were segmented simultaneously. (best viewed in color)

### *F. Computational efficiency*

We implemented a non-optimized version of FOMTrace in C/C++, providing a user interface that allows the visualization of the result on-the-fly and interactive intervention when necessary. The method goes through a pre-processing stage in which the superpixel-graph is created and optical flow is computed using the Matlab code from [31]. We obtained an upper bound for FOMTrace's computational time by applying it without user intervention to video `bird_of_paradise` (Figure 5), in a machine with a 3.5 GHz Intel Core i7 and 32 GB of RAM. The video has length $n_f = 98$ with $640 \times 360$ pixels frames. The object occupies a large portion of each one.[5] The pre-processing step takes about 5.2s per frame, and the interactive stage consumes about 1.8s per frame, linearly decreasing towards the end of the video. In FOMTrace$^{\mathrm{W}}$, the computation of weight image $W_t$ is an expensive step, which further adds another 2.2s per frame to the interactive stage.

In general, we do not recommend using the entire video to build the superpixel-graph. Although having optimum-paths coming backwards in time is helpful for label prediction, they may also harm segmentation when complex leakings occur in the future. We thus give the user the option of running the method for a variable length portion of the video, which is usually 30 frames long. Hence, the interactive time decreases to about 0.6s for FOMTrace$^{\mathrm{W}}$ and 2.8s for FOMTrace$^{\mathrm{W}}$. We may replace our weight computation via inpainting by another simpler, faster technique in the future, similarly to [11].

## IV. Conclusion and Future Work

We have presented FOMTrace, an interactive video segmentation method that relies on two types of image graphs and fuzzy object models. FOMTrace first automatically segments the video object via optimum seed competition from an input labeled frame, using IFT on the superpixel graph. Then, our method derives a fuzzy image from the predicted and past labels to refine the automatic segmentation result on a pixel graph for the current frame, also using IFT-SC. The user verifies if the segmentation result is as desired and may correct it using IFT's integrated framework [7], [15], [43], [49]. FOMTrace uses the final label to restart the process in the next frame, and iterates until the end of the video.

Our experiments have shown FOMTrace to be competitive with the state-of-the-art in interactive video segmentation, supervised and unsupervised object tracking. Moreover, we have designed FOMTrace as a flexible method with building blocks that may be improved with several ideas. For instance, the weighted computation of fuzzy image $O_t^W$ in the Appendix imposes a hard constraint for FOMTrace$^{\mathrm{W}}$ that must be used with caution. We need to soften it since it increases the method's dependence on optical flow. Likewise, determining the frames in which the fuzzy image $O_t$ should be used at all to refine segmentation is an interesting topic of study.

In the future, we will also investigate the use of temporal superpixels as described by Chang and Fischer [51] to build our graphs. We also plan on incorporating hierarchical long-term supervoxel cues, as was done in [32], in our graph's arc weight estimation. This promises to improve the automatic video segmentation step using IFT-SC.

Experiment-wise, it would be interesting to measure the amount of user effort and time taken to correct video segmentation, following [8], for other competing methods. In the future, we aim to use the recently proposed open source segmentation tool named SENSAREA [52] as a basis for comparison with our methods. We note that the work of Gallego and Bertolino [14], used for indirect comparison here, is planned to be added to SENSAREA in the future [52].

[5]Although the IFT algorithm may run in time $O(|\mathcal{N}|)$, our version for superpixels runs in $O(|\mathcal{N}| \log |\mathcal{N}|)$ regardless of the object's size [17]. The object's size only matters for inpainting-based weight image computation [50].

## APPENDIX

By setting all pixels in $\widetilde{W}_t$ with weight 0.5, we strike a balance between the propagated and predicted label images in Eq. 2. This is particularly important for fast-moving deformable objects, since the predicted label $\hat{L}_t$ tends to better delineate them in the current frame $I_t$, when foreground and background have mostly distinct colors.

Even though the balanced weight image is useful to stop leakings on foreground regions where the touching background has similar colors (e.g., Figure 8), it does not always suffice (Figure 14). In those regions, the shape constraint provided by the propagated label $\widetilde{L}_t$ should be given greater importance to ensure that foreground and background model seeds from $O_t$ be selected around them, similarly to the work in [9].

We intend to determine, in the previous frame $I_{t-1}$, the foreground regions that are most similar to the surrounding background to compute the weight image $\widetilde{W}_t$. We accomplish this task by computing an absolute difference image $J_{t-1}(p) = \left\| I_{t-1}(p) - \tilde{I}_{t-1}(p) \right\|_1$ normalized between $[0.0, 1.0]$. $\tilde{I}_{t-1}$ is the result of inpainting the object mask from the final label $L_{t-1}$ in frame $I_{t-1}$ [50].[6] Then, we compute a weight image $W_{t-1}(p) = 1.0 - J_{t-1}(p)$.

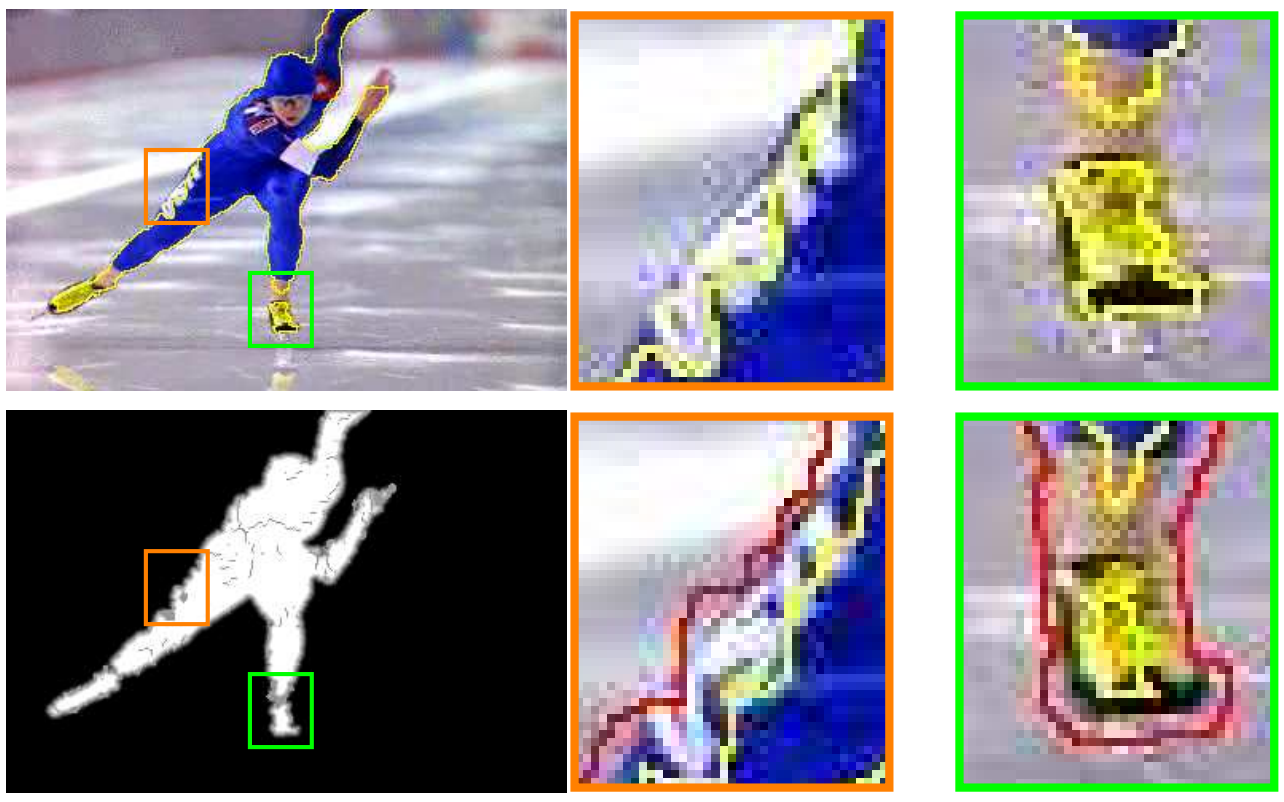

Fig. 14: **Top row:** segmentation leaks of mask $L'_t$ on regions where foreground and background share similar colors and touch each other, which occurred due to setting 0.5 as the weight of all pixels in image $\widetilde{W}_t$ (Eq. 2). **Bottom row:** corresponding fuzzy object model $O_t$ and part of its foreground/background seeds that improperly surround the problematic regions.

Intuitively, inpainted background regions from $\tilde{I}_{t-1}$ that have similar colors to the object appearing in $I_{t-1}$ will be given higher scores in $W_{t-1}$, indicating possible locations where leaking may occur. We then create $\widetilde{W}_t$ by propagating $W_{t-1}$ via optical flow as is done for $\widetilde{L}_t$ (Figure 15a).

Finally, to ensure that seeds be tightly selected around weaker parts of the object's boundary (Figures 15b-15c), we compute $O_t$ conservatively by forcing our method to consider *only* the propagated label's sEDT if the weight for a pixel $p \in D_{\widetilde{W}_t}$ is above a user-controlled threshold $\gamma$:

$$O_t^W(p) = \begin{cases} \widetilde{E}_t(p) & \text{if } \widetilde{W}_t(p) \geqslant \gamma, \\ O_t(p) & \text{otherwise,} \end{cases} \quad (3)$$

where $O_t(p)$ is computed as in Eq. 2. Fuzzy model image $O_t^W$ in Eq. 3 may be used instead of $O_t$ from Eq. 2 to produce a more refined segmentation label in those cases (Figure 15d).

[6]For multiple objects, we binarize $L_{t-1}$ when inpainting frame $I_{t-1}$.

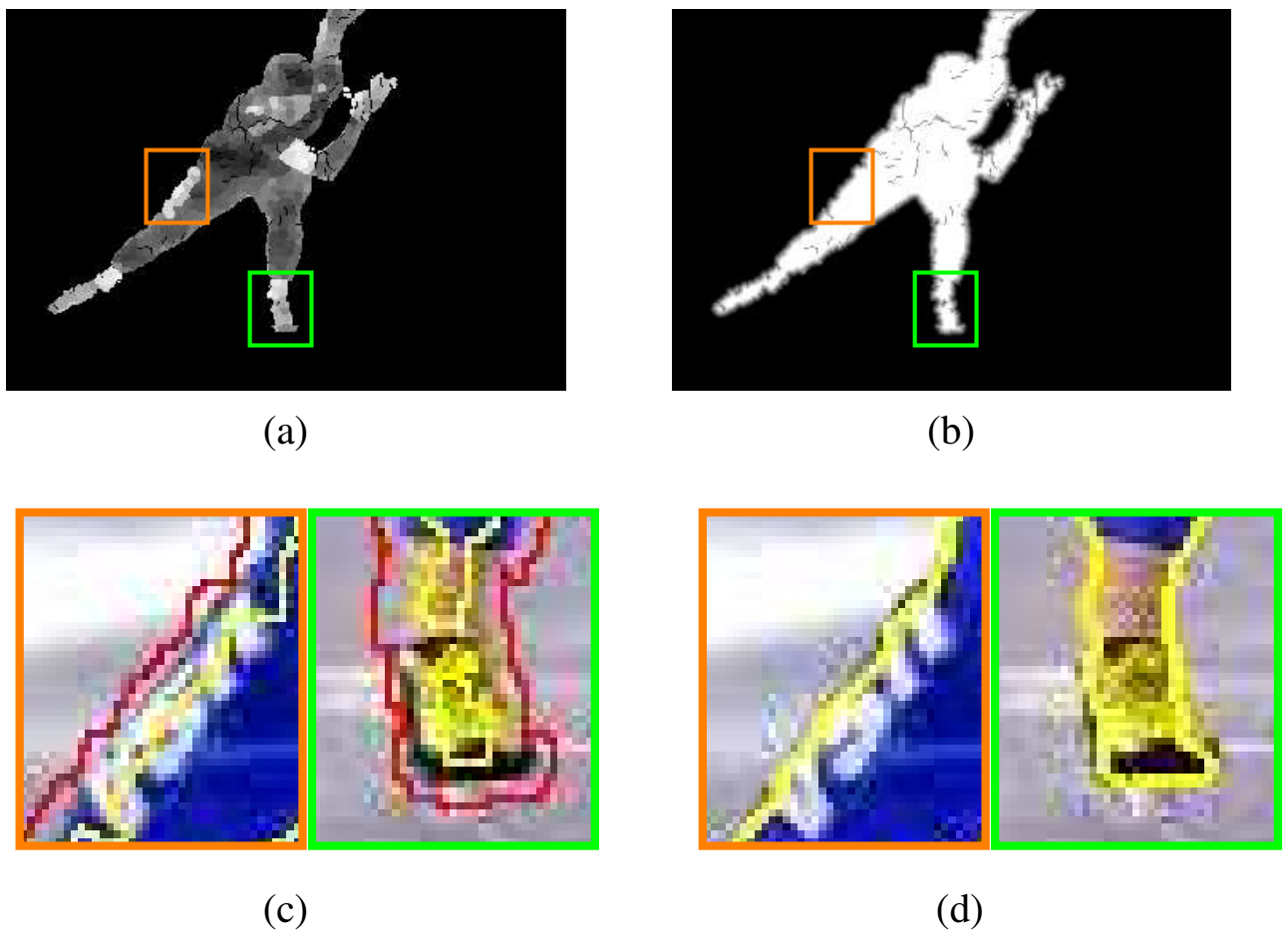

Fig. 15: (a) Propagated weight image $\widetilde{W}_t$. (b) Fuzzy object image $O_t^W$ computed according to Eq. 3. (c) FOM seeds tightly selected around weaker parts of the object's boundary. (d) Refined segmentation result for label image $L'_t$.